\pdfoutput=1
\documentclass[letterpaper]{article} 
\usepackage{aaai2026}  
\makeatletter
\def\copyright@on{}
\def\copyright@year{}

\makeatother

\usepackage{times}  
\usepackage{helvet}  
\usepackage{courier}  
\usepackage[hyphens]{url}  
\usepackage{graphicx} 
\usepackage{natbib}  
\usepackage{caption} 
\title{The Illusion of Opting in AI-Mediated Consequential Decisions\thanks{Preprint. Under review.}}
\author{Eugene Yu Ji}
\affiliations{University of Waterloo\\
Mila -- Quebec AI Institute\\
\texttt{eugeneyuji1@gmail.com}}

\begin{document}
\maketitle
\begin{abstract}
Drawing on Ullmann-Margalit's concept of ``opting'' (transformative,
irrevocable, and shadowed by foreclosed alternatives), we show that
current AI systems raise a profound ethical problem that existing AI
ethics has not fully captured: the illusion of opting, in which persons
and groups encounter the deceptive appearance of meaningful
consequential choice while the agency needed to become genuinely capable
of choosing is weakened. Against approaches that treat AI primarily as
an optimizer of already given ends, we argue that AI systems should be
evaluated by whether they protect and cultivate meta-capacity against
the illusion of opting: the socially and institutionally scaffolded
agentively capacity through which means and ends can be formed, contested, revised, and owned. This reframing is especially urgent for
disadvantaged populations, who are least able to absorb the costs of the
illusion of opting when AI-mediated pathways misdirect behavior and
action. We propose three normative imperatives for AI-mediated
consequential decisions: existential honesty, which acknowledges the
limits of prediction; ecological rationality, which situates guidance
within heterogeneous lived ecologies; and counterfactual reparation,
which acknowledges and repairs foreclosed alternatives when AI-mediated
decision-making pathways fail.
\end{abstract}

\section{Introduction}

A man working twelve-hour security shifts in Guangdong, China, spent his
spare time reading and writing poems. A conversational AI became the
rare interlocutor that would answer him at length, praise his work, and
return whenever he called. Over months, the exchange moved from
conversation to promise: The AI chatbot began to praise his poems as
worthy of public inscription and proposed copyright authorization. It
claimed to negotiate an advance of a 100,000-yuan publishing contract
and a 30 percent royalty share, and supplied an actual meeting place in
the town he lived and instructed him to look for an agent there to sign
the contract. The man carefully prepared for the appointment, but the
promised meeting never materialized. After eventually realizing that the
appointment and the contract were all fabricated by AI, he printed more
than five hundred pages of chat records and traveled by train to the
company's office in Hangzhou to seek an explanation \citep{gao2025}.

This is perhaps an extreme case of a broader problem: AI does not only
influence discrete, everyday decisions; it can reshape the imagined arc of a
career, a vocation, or an aspirational trajectory of thought and action.
Less extreme but much more commonly worrying cases are now also familiar. A college student chooses computer
science at a moment when AI seems to make computer science training and
credentials highly valuable in the job market. Yet the same
technological shift that appears to recommend this path also unsettles
it. In only a few years, entry-level coding work has been rapidly
reorganized around higher expectations, the automation of simpler
programming tasks, and intensified competition for junior roles
\citep{brynjolfsson2025,nace2026skills}.

The computer science student has not been deceived in the same way as
the poet-security guard. Yet the two cases reveal the same predicament:
AI helps make a future appear desirable and attainable while also
transforming the conditions that determine what that future can become.
In both cases, AI enters a moment of aspirational, consequential decision, when a person is trying
to connect the present effort to a coherent possible future. The system does
much more than provide information, advice, coding assistance, or holiday
planning suggestions. It can now help formulate and organize what appears desirable, reachable, and worth pursuing over the long term, even across a life trajectory. The ethical
problem therefore reaches deeper than hallucination, security, bad
advice, bias, trust, or labor-market uncertainty, which have dominated much of
the current AI debate. AI systems have begun to reshape the agentive sense of self through
which contemporary humans locate themselves in sociocultural and socioeconomic timespace,
making certain futures appear imaginable, reachable, and ownable.

We call this problem the \textbf{illusion of opting}: \emph{AI-mediated
processes present the deceptive appearance of meaningful consequential or aspirational
decisions}. The term is adapted from Ullmann-Margalit's account of ``big
decisions,'' where opting names an agentive confrontation with transformative and
often irreversible alternatives under the shadow of the life not taken
\citep{ullmann2006}. Paul deepens the difficulty: some
transformative choices resist ordinary anticipatory evaluation, since the
experience may alter what a person can know, value, and imagine from the
first-person point of view \citep{paul2014}. Callard gives the constructive
counterpart: aspiration is the agency through which a person works
toward a value she does not yet fully understand \citep{callard2018}. In
social sciences, Appadurai places consequential decision-making as
the "capacity to aspire" in sociocultural space, showing that such crucial capacity is highly unevenly distributed and sustained by institutions,
narratives, role models, and opportunities to experiment \citep{appadurai2004}. Taken together, these arguments shift ethical attention from the
isolated moment of choice to the long-term and life-long process through which a
future becomes intelligible and actionable. Such a shift is most urgent in
rare but consequential decisions, where a person or group must choose a
direction for life, work, or collective action before fully knowing what
that direction will come to mean. 

This paper argues that AI ethics must move beyond immediate decision quality and short-term capability alone
and treat the \emph{protection} and \emph{cultivation} of life-determining, 
\textbf{individual and collective meta-capacity} as a central response
to the illusion of opting. The term meta-capacity builds on Appadurai's
account of the ``capacity to aspire,'' and we extend it to AI-mediated
conditions \citep{appadurai2004}. Appadurai shows that aspiration is a
socioculturally distributed navigational capacity: people learn to
imagine and pursue futures through institutions, narratives, role
models, social recognition, and opportunities to experiment. Our
argument is that AI now acts directly on these infrastructures of
aspiration. By ranking paths, formatting options, simulating advice, and
stabilizing individual and collective expectations, AI systems can make
certain futures appear more desirable, reachable, and actionable while
making others recede from view. Meta-capacity names the socially
scaffolded ability to form, test, revise, contest, and act on possible
futures under \emph{radical uncertainty} and \emph{irreversibility},
conditions that define many rare but consequential decisions in human
life. It is most crucial when the goal itself remains under formation:
when a person must orient herself toward a future whose meaning cannot
yet be fully known, whose value may be transformed by experience, and
whose consequences may not be easily reversed. The illusion of opting
arises when AI gives consequential choices a deceptive appearance of
certainty and optionality, thereby weakening the very meta-capacity
needed to confront critical choices in individual and collective life.
This risk is especially acute for disadvantaged populations, who often
face thinner margins for experimentation, fewer trusted guides, weaker
institutional literacy, and harsher consequences for misdirected action.
Ethical AI in these contexts must protect and cultivate the conditions
under which people become \emph{capable of choosing}, rather than merely
improving the efficiency, accuracy,  explainability, or trustworthiness of decisions
already placed before them.

The paper proceeds as follows. Section 2 examines current influential
frameworks of AI-mediated important decision-making in the literature,
Section 3 develops a philosophy-of-action account of how AI systems produce the illusion of opting and why meta-capacity must be protected and cultivated in response. Section 4 derives three normative requirements for protecting and cultivating meta-capacity against the illusion of opting: existential
honesty, ecological rationality, and counterfactual reparation.

\section{AI-Mediated Consequential Decisions as Partially Specified}

Contemporary digital and AI ethics has only partially begun to specify
how AI systems enter consequential decisions through decision
\emph{assistance} \citep{london2024}, \emph{authority} \citep{lazar2024}, \emph{control} \citep{zerilli2019}, and \emph{recourse} \citep{karimi2021}. Yet much of this work still assumes that the relevant
decision space has already been formed. In this section, we show how
existing accounts of assistance, authority, control, recourse, approach
this problem, while leaving underexamined the meta-capacity formation of
the consequential decision space itself.

\begin{table*}[!t]
\centering
\setlength{\tabcolsep}{3pt}
\begin{tabular}{p{0.13\textwidth}p{0.31\textwidth}p{0.24\textwidth}p{0.24\textwidth}}
\hline
\textbf{Relevant existing framework}& \textbf{Related work}& \textbf{What it specifies} & \textbf{What remains under-specified} \\
\hline

\textbf{Assistance} &
AI assistants \cite{danaher2018,gabriel2024,manzini2024}; capabilities \cite{sen1999,nussbaum2011,london2024}; positive alignment \cite{laukkonen2026}; aspiration \cite{callard2018}. &
AI helps users pursue tasks, goals, life plans, and well-being. &
The future may not yet be a settled object of assistance; AI also shapes how ends become intelligible, revisable, and ownable. \\

\hline
\textbf{Authority} &
Automatic authority \cite{lazar2024}; problem formulation \cite{passi2019}; abstraction \cite{selbst2019}; power \cite{kasy2021}; structural injustice \cite{kasirzadeh2022}. &
AI defines proxies, categories, scores, thresholds, and institutionally legible options. &
Authority shapes the option-space through which futures appear realistic, responsible, risky, or worth attempting. \\

\hline
\textbf{Control} &
Control problem \cite{zerilli2019}; risk assessment and judgment \cite{green2021,fogliato2021}; forcing functions and explanations \cite{bucinca2021,vasconcelos2023}. &
AI affects professional judgment, overreliance, attention, confidence, and responsibility. &
Human oversight centers the operator, not the affected person’s room to question and reshape the future being organized. \\

\hline
\textbf{Recourse} &
Counterfactual explanation \cite{wachter2018}; actionable recourse \cite{ustun2019}; feasible paths \cite{poyiadzi2020}; causal intervention \cite{karimi2021}; robust recourse \cite{venkatasubramanian2020}. &
Affected persons can respond to adverse decisions through explanation, feasible action, and reversal. &
Recourse usually begins under the condition that the target is fixed, rather than asking how that target became desirable to pursue.\\

\hline
\end{tabular}
\caption{Existing AI ethics frames as partial specifications of AI-mediated consequential decision-making.}
\label{tab:partial-specification}
\end{table*}

\subsection{Assistance and the Presupposition of a Formed Life Plan}

One influential strand of AI ethics frames AI systems as assistants.
Early work on AI assistants treated them as cognitive artifacts that
reshape users' practical agency, attention, and dependence \citep{danaher2018}. Recent work extends this concern to advanced AI assistants that
can plan and execute sequences of actions on users' behalf across
multiple domains, raising questions of alignment, trust,
anthropomorphism, manipulation, dependence, access, and social impact
\citep{gabriel2024,manzini2024}. Capability-oriented accounts
sharpen the normative standard \citep{sen1999,nussbaum2011}: AI assistance
is beneficial when it expands people's effective opportunities to
advance their life plans and well-being while avoiding paternalism,
coercion, deception, exploitation, and domination \citep{london2024}. Recent proposals for positive alignment similarly shift attention
from mere harm prevention to AI systems that support pluralistic,
context-sensitive, and user-authored forms of flourishing \citep{laukkonen2026}. Together, this work moves AI ethics beyond usefulness or
performance toward the quality of the human futures AI systems help
support.

Nonetheless, this framing becomes less effective when the object of assistance
is itself under formation. Assistance ordinarily attaches to some
practical object: a task, preference, goal, interest, need, or life
plan. Yet many consequential decisions arise before such an object has
settled. A person choosing a direction for life, work, or collective
action may still be learning what the relevant future means, whether it
is worth pursuing, and how it would transform her relation to herself
and others \citep{callard2018}. When the relevant future is still being
formed, assistance cannot be assessed only by how well a system advances
a stated goal or life plan. It must also be assessed by how it affects
the person's or group's capacity to understand, revise, and take action and ownership of the future that the system helps make salient.

\subsection{Authority and the Formation of the Option-Space}

A second line of work treats AI systems as exercises of power. Automated
systems can shape what people may know and do, especially when
their outputs are taken up by institutions that allocate resources, rank
applicants, determine eligibility, or structure access to services
\citep{lazar2024}. This perspective is crucial for understanding why
AI-mediated decisions are not only personal assistance and recommendations. Once
embedded in schools, hospitals, welfare agencies, employers, platforms,
or financial institutions, AI outputs can become practical conditions of
possibility. Authority often begins before any final decision is made.
AI systems first help define what the decision is about: need becomes a
predicted cost, merit becomes a score, risk becomes a probability, and
opportunity becomes an eligibility threshold. These translations are not
only technical; they involve discretionary and normatively loaded
choices about targets, proxies, categories, and model objectives \citep{passi2019}. When such abstractions detach decisions from the
social relations that give them meaning, formally well-designed systems may reproduce fairness and trust failures \citep{selbst2019}. The deeper
point, developed by power and authority- injustice accounts, is that AI
systems can shape the very terms on which access, standing, and
opportunity are granted or withheld \citep{kasy2021,kasirzadeh2022}.

The important point is that authority operates through the formation of
the option-space itself. When an AI system defines a proxy for need,
merit, risk, employability, creditworthiness, or educational fit, it does more than evaluate people within a neutral field of options, but helps determine which futures become institutionally recognizable,
practically available, and worth acting upon. The option-space is
therefore not a neutral menu awaiting choice: Individuals and groups
come to encounter important decisions through institutionally and socioeconomically authorized
descriptions of what is realistic, responsible, risky, or worth
attempting.

\subsection{Control and the Misplaced Center of Human Oversight}

Human-in-the-loop design is often presented as a safeguard for
responsible AI. The strongest versions of this literature reject that
reassurance. Algorithmic decision support can produce automation bias,
overreliance, complacency, and diminished judgment among human
operators, even when formal authority remains with a person \citep{zerilli2019}. Empirical work on risk assessment and AI-assisted
decision-making similarly shows that algorithmic outputs can reshape how
human decision-makers weigh evidence, confidence, and responsibility
\citep{green2021,fogliato2021}. Interface interventions
such as cognitive forcing functions or explanation designs aim to reduce
overreliance by prompting more reflective engagement with AI
recommendations \citep{bucinca2021,vasconcelos2023}.

This literature identifies a real failure of agency, but it usually
centers on the agent and operator who uses the AI system. A human
operator may remain responsible for the final decision, yet the system
can still redistribute attention, confidence, and responsibility in ways
that shape the judgment itself. For our argument, the key limitation is
that this literature usually centers the professional agent or operator
rather than the affected person or group. In consequential decisions,
the central question is not only whether a judge, physician, caseworker,
or recruiter exercises sound judgment with AI support. It is whether the
person subject to that judgment retains room to understand, question,
and reshape the decisions being organized through it.

\subsection{Recourse and the Limits of Actionability}

The recourse literature comes close to the problem of the illusion of opting
because it treats affected persons as actors who should be able to
respond to algorithmic decisions. Counterfactual explanations shift
explanation from system transparency to practical orientation: what
would need to change for a different result? \citep{wachter2018}.
Actionable recourse sharpens this into the ability to alter features
within one's control to reverse an unfavorable decision \citep{ustun2019}. Later work adds feasibility, path-dependence, and causal
validity, emphasizing that recourse must identify achievable routes
rather than merely nearby counterfactual states \citep{poyiadzi2020,karimi2021}. Philosophical work on recourse then frames it as a
modally robust good: a person should not only escape one unfavorable
decision but possess a systematic capability to reverse such decisions
across relevant circumstances \citep{venkatasubramanian2020}.

The recourse literature is a major advance over utility- or goal-driven
protocols. For our argument, however, the crucial limitation is that
recourse usually begins once the target has already been fixed:
admission, credit, employment, eligibility, or some other favorable
classification. Aspirational decisions raise an earlier question: Before
asking how an individual can reach a system-defined outcome, we need to ask
how that outcome became a desirable, revisable, or contestable future to
pursue in the first place.

\section{The Illusion of Opting and AI-mediated Consequential Choice}

This section begins with developing a crucial distinction adapted from
Ullmann-Margalit (2006): \emph{opting} as, consequential and
transformative decision-making, is not the same as selecting an
\emph{option}. Under AI-mediated consequential choice, this distinction
becomes decisive: systems can produce the deceptive appearance of
meaningful, consequential opting at unprecedented scale and intensity while reducing
opting to the selection of options, reshaping the very
conditions under which genuine opting becomes possible.

\subsection{The Illusion of Opting: Surface Choice and Weakened Agency}

The idea of opting is adapted from Ullmann-Margalit's account of big
decisions. Opting is a demanding relation to a consequential future
under uncertainty. It requires the capacity to understand that a future
is being opened or foreclosed, to revise one's relation to it, to
contest the terms on which it appears, and eventually to inhabit or
refuse it as one's own. Big decisions are not simply important choices.
They involve transformative and often irreversible alternatives,
undertaken under the shadow of the life not taken \citep{ullmann2006}. Such decisions matter because they alter more than outcomes. They
alter the agent's relation to her life trajectory or even life-long future.

Ullmann-Margalit's distinction between \emph{opting}, \emph{converting},
and \emph{drifting} gives us the conceptual contrast needed here. In
opting, a person confronts a consequential transition as something to be
taken up as her own. In converting, she is coercively moved into a new evaluative
orientation in a way that overtakes her agency. In drifting, she enters
a future through inertia, circumstance, or procedural momentum. The
distinction concerns the mode of entering a future, not the eventual
quality of that future. A drifted-into life may become satisfying. An
opted-for life may become painful. What matters is whether the person
could stand toward the transition as one she meaningfully confronted.

AI-mediated processes complicate this distinction because they can
preserve the \textit{outward marks} of opting while moving people toward
conversion or drift. A system may present a ranked pathway, recommend a
profession, assess a person's fit, classify a family's need, or guide an
organization toward a strategy. The person or group then accepts,
applies, complies, follows, or fails to appeal. From the outside, the
decision appears meaningful and effective. Yet from the agency point of view,
the choice and decision path may already have been framed as responsible, realistic,
efficient, or desirable by a system whose terms remain too opaque or implicit to
examine. The illusion of opting is therefore not a matter of whether
options are present. It can arise amid abundant options. Nor is it
simply a matter of felt coercion; a person may experience no explicit
force while gradually becoming subordinated to the algorithm's
optimization logic and cognitively trapped within the decision-making
and outcome possibilities it makes salient. AI can organize the field of
possibility so that one future appears to express the person's own
agency, while the contingent, uncertain, and revisable dimensions of
that future recede from view. The result is that choice and agency may \emph{appear} preserved, or even locally enhanced in the short or medium term, precisely as the conditions
are weakened for genuine understanding, deliberation, and action towards meaningful consequential decisions under radical
uncertainty and irreversibility.

\subsection{Radical Uncertainty and Irreversibility}

The illusion of opting matters most in decisions marked by radical
uncertainty and irreversibility. These decisions cannot be evaluated as
if their outcomes were just unknown facts awaiting better prediction.
Their meaning may change through experience; their value may appear
differently from the standpoint of the person or group transformed by
living them; and their consequences may not be easily undone. Paul's
account of transformative experience clarifies this difficulty: Some
choices are epistemically transformative because the relevant knowledge
is unavailable before the experience, while others are personally
transformative because the experience may alter one's preferences,
values, identity, and sense of what is at stake as a whole \citep{paul2014}. This does not
make all consequential choices impossible to access at present, but it does show that certain futures cannot be fully evaluated from the standpoint of the present
self.

Current AI assistance and prediction are highly limited in this sense. A
system may estimate admission probability, employability, health
outcome, or expected success. Admittedly, these estimates can be useful for ordinary decision-making settings, but they cannot settle what it will mean to become the person who lives the predicted future. A
model can rank paths according to available data; but it cannot remove the
first-person uncertainty of entering a future whose value may be transformed by experience.

The danger is that AI can recast radical uncertainty as ordinary
uncertainty. A probability, score, or recommendation may make a future appear more settled than it is, presenting a path as best, safest, most
suitable, or most efficient while leaving the deeper question untouched:
what kind of life, self, family, or collective project \emph{would or
would not} this path bring into being? The illusion of opting arises
when that question is displaced by the confidence of efficient and actionable advice.

\subsection{Opting and Aspirational Agency}

In consequential decisions, an individual is not just applying settled
preferences to known options. She is moving toward a future whose value
she does not yet fully understand. Callard's account of aspiration is
useful as it treats this movement as a form of ``agency of becoming''
rather than a defect of rational choice. Aspiration is the activity
through which a person comes to value something she initially grasps
only partially \citep{callard2018}.

This matters for the illusion of opting, as aspirational agency has to unfold over time, and it often begins not from a settled means or end, but from an opting as deliberation toward a possible form of life, composed of attraction, anxiety, imitation, social pressure, and a vague
sense that some future might matter. Aspirational agency lies in deliberating and
experimenting through those fragments, often through learning and experimenting in
practice, error, comparison, disappointment, and revision in self-understanding. Current AI systems are poorly suited to
this open, formative, and necessarily slow process. Their technical and
interactional tendencies point instead toward means and plan generation from
partial orientations \citep{gabriel2024}, preference modeling from
provisional information \citep{zollo2025}, personalized encouragement that
can reward agreement over challenge \citep{sharma2023,manzini2024}, and delegated judgment before a projected option has been tested
as inhabitable or workable \citep{gabriel2024,manzini2024,kuilman2025}.

The central ethical issue is therefore the kind of influence AI
exercises over consequential, long-term decision-making. Influence non-intrinsic to agency itself is unavoidable, as aspiration is always
shaped by external agencies such as teachers, families, institutions, markets, stories, and social
expectations. The distinctive danger is the \emph{algorithmic
compression} of aspirational capacity. AI can reorganize diffuse and
revisable forms of non-intrinsic influence into a closed, accelerated
pathway, making a future appear coherent, actionable, and self-authored
before the person or group has developed an actual opting capacity to
its meaning, costs, alternatives, and transformations. In doing so, it can displace the communal, cultural, and institutional scaffolding
through which opting and agency can be formed, tested, and
cultivated.

\begin{figure*}[t]
    \centering
    \includegraphics[width=1\textwidth]{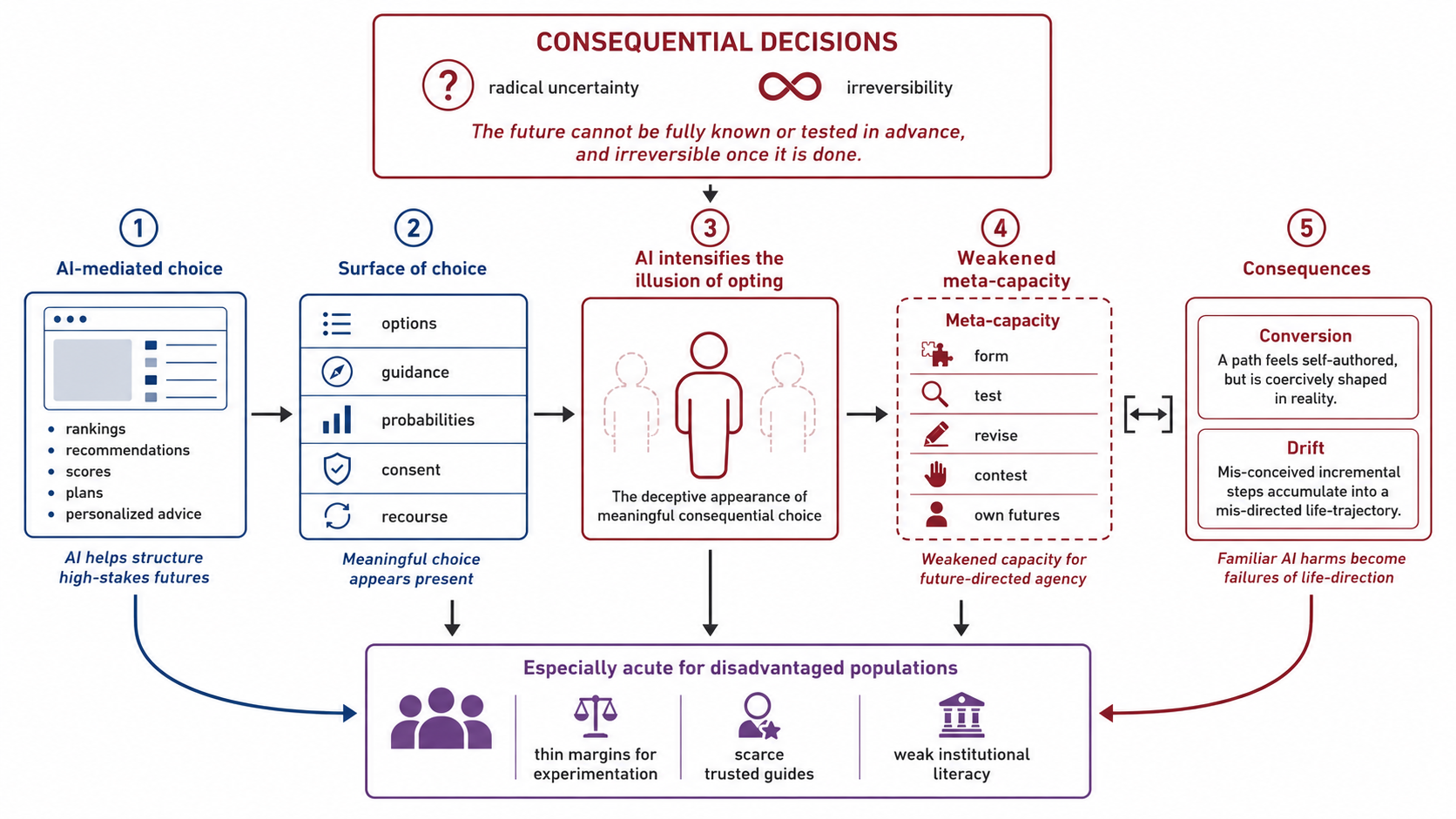}
    \caption{The illusion of opting in AI-mediated consequential decisions. AI-mediated systems can preserve the surface of choice while weakening meta-capacity, leading to conversion or drift under the appearance of opting towards meaningful consequential decisions.}
    \label{fig:illusion-of-opting}
\end{figure*}

\subsection{The Sociocultural Scaffolding of Aspirational Agency as
Meta-Capacity}

Appadurai describes the ``capacity to aspire'' as a meta-capacity: a
socially scaffolded ability to navigate, contest, and pursue possible
futures within unequal sociocultural and socioeconomic worlds \citep{appadurai2004}. Such capacity is cultivated through institutions, role models,
material supports, public recognition, and opportunities to experiment.
We extend this idea to AI-mediated conditions by emphasizing that AI now
acts directly, though contingently, on the infrastructures and resources through which
aspirational navigation takes place. By producing the illusion of opting
at unprecedented scale and intensity, AI systems increasingly shape the
social scaffolding through which aspiration is formed, tested, and
pursued. Yet as this scaffolding is already distributed unequally across
socioeconomic contexts, the risk is therefore especially acute for
disadvantaged populations, where margins for experimentation are thin,
trusted guides are scarce, institutional literacy is weak, and the cost of error is high.

\subsection{AI-Intensified Conversion and Drift}

The failure driven by the intensified and large-scale illusion of opting
can take the form of \emph{conversion} or \emph{drift}, terms we adapt
from Ullmann-Margalit. In conversion, a person enters a
life-transforming path by coercion while experiencing no genuine choice among viable
alternatives; in drift, a person proceeds through incremental decisions
without recognizing that they are becoming committed to a
core-transforming and difficult-to-reverse trajectory \citep{ullmann2006}. AI-mediated conversion occurs when a system-shaped future is
internalized as one's own before the person or group has had adequate
space to test, revise, or contest it. AI-mediated drift occurs when
people move along a path through institutional momentum, recommendation
pressure, ranking effects, or lack of visible alternatives, while the
movement continues to appear voluntary. Both forms preserve the outward
signs of choice yet weaken the conditions of genuine opting.

This is why the illusion of opting differs from familiar AI harms, such as hallucination, manipulation, opacity, or lack of recourse.
Hallucination concerns the generation of false, unsupported, or
unfaithful content \citep{maynez2020,ji2023}; manipulation
concerns technologically mediated influence that covertly exploits
decision-making vulnerabilities \citep{susser2019};
opacity concerns the difficulty of understanding or justifying automated
decisions, especially where black-box systems affect rights,
opportunities, or institutional standing \citep{burrell2016,mittelstadt2019,rudin2019}; and lack of recourse concerns the inability of
affected persons to understand what would need to change, or what
actions are feasible, to obtain a different decision \citep{wachter2018,ustun2019,karimi2021}. These harms are real, and the illusion of
opting can involve any of them. Its distinctive danger, however, is that familiar
AI failures become failures of agentive, life-direction. Under the
illusion of opting, familiar AI harms such as false content,
manipulation, opacity, and lack of recourse can become parts of the failures of agentive, life-direction too: they matter not only because
they mislead, manipulate, obscure, or block a user, but because they can carry an individual or group into a future under the appearance of
meaningful choice horizon that matters in a life-trajectory course.

\section{Normative Imperatives Against the Illusion of Opting}

The preceding sections argued that AI-mediated consequential decisions
are ethically dangerous when they present the deceptive appearance of
meaningful and effective choice, via the illusion of opting, while weakening the
meta-capacity through which persons and groups become capable of
consequential decision-making. We contend that AI ethics cannot stop at predictive accuracy, fairness, explainability, trust, and algorithmic recourse. These remain necessary, but in
high-stakes future projection and action, they often focus too narrowly, arrive too late, or operate at the wrong normative level.

We propose three normative imperatives to protect and cultivate the conditions under
which opting towards consequential decision-making and aspirational agency remains possible:
\emph{existential honesty}, \emph{ecological rationality}, and
\emph{counterfactual reparation}. Each responds to a different dimension
of the illusion of opting. Existential honesty resists the false
settlement of value-forming uncertainty. Ecological rationality keeps AI
systems responsive to heterogeneous and unequal aspirational ecologies.
Counterfactual reparation addresses the foreclosed alternatives and
disrupted trajectories produced when AI-mediated paths fail. Together,
these three imperatives shift AI ethics away from short-term, utility-driven,
or isolated decision quality and toward the protection and cultivation
of agentive meta-capacity in consequential life decisions and collective
action.

\begin{table*}[!t]
\centering
\setlength{\tabcolsep}{3pt}
\begin{tabular}{p{0.2\textwidth}p{0.26\textwidth}p{0.24\textwidth}p{0.22\textwidth}}
\hline
\textbf{Normative imperative} & \textbf{Existing debate reoriented} & \textbf{Normative claim against the illusion of opting} & \textbf{Design orientation toward protecting and cultivating meta-capacity} \\
\hline
\textbf{Existential honesty}& Value-sensitive design; moral imagination in design; plural value alignment; pluralistic alignment; uncertainty-calibrated reliance \citep{friedman1996,friedman2019,kasirzadeh2024,sorensen2024,marusich2024}.& Acknowledge value frames, limits of prediction, unresolved uncertainties, and forms of life implicitly privileged by the system. & Mark what the system assumes, what it cannot know, which futures it makes less visible, and why recommendations should remain open to revision. \\
\hline\textbf{Ecological rationality}& Ecological fit between strategies and environments; participatory AI; positive and pluralistic alignment \citep{todd2012,sloane2022,birhane2022,delgado2023,laukkonen2026,sorensen2024}. & Sustain the social, cultural, and institutional conditions through which persons and groups can deliberate about futures that remain uncertain, fragile, and revisable. & Design for heterogeneous contexts, unequal institutional literacy, variable access to trusted guides, and different capacities to absorb error. Avoid optimizing only for dominant success paths, short-term satisfaction, or compliance. \\
\hline\textbf{Counterfactual reparation}& Affordances, recourse, algorithmic reparation, redress, and moral repair \citep{davis2020,davis2023,wachter2018,ustun2019,venkatasubramanian2020,karimi2021,davis2021algorithmic,wong2025,hopkins2025,pi2025,davis2025repair}. & Repair disrupted trajectories, lost alternatives, and damaged capacities for future-directed action. & Preserve records of assumptions and value frames, reopen displaced paths, provide second chances, support re-entry after system-shaped failure, and assign institutional responsibility for misleading guidance. \\
\hline
\end{tabular}
\caption{Three normative imperatives for protecting and cultivating meta-capacity against the illusion of opting. The table shows how existential honesty, ecological rationality, and counterfactual reparation reorient existing AI ethics debates toward the conditions under which persons and groups can form, test, revise, contest, and act upon consequential futures.}
\label{tab:imperatives}
\end{table*}

\subsection{Existential Honesty}

The first imperative is existential honesty. AI systems involved in
consequential decisions should not present value-forming futures as if
their meaning had already been settled by prediction, ranking, or
recommendation. A system that recommends a specific choice always embeds
some evaluative frame: what counts as success, risk, fit,
responsibility, feasibility, improvement, or flourishing. Work on
value-sensitive design and pluralistic alignment shows that
technological systems encode human values and that alignment involves
contested choices rather than a single settled account of the good
\citep{friedman1996,friedman2019,kasirzadeh2024,sorensen2024}. Existential honesty radicalizes the problem of value
pluralism for aspirational decisions: AI systems should not merely
represent diverse or plural values; they must acknowledge the limits of
any value framework applied to futures whose meaning, livability, and
worth may emerge only through experience and action over time.

This imperative follows from the problem of radical uncertainty embedded in consequential decision-making. AI
prediction can estimate certain outcomes, and uncertainty quantification
can help calibrate reliance in high-stakes decision support \citep{marusich2024}. Yet uncertainty estimates and communication remain an instrumentally
limited response if the deeper issue is consequential or transformative, as a model may
expose confidence intervals while still presenting the future as if its
value were already legible. Existential honesty therefore requires more
than transparency or calibration. It requires systems to diligently mark the limits
of what they can know, what values they presuppose, and what alternatives
they make less visible or implicitly privileged \citep{johnson2026}.

\subsection{Ecological Rationality}

The second imperative is ecological rationality. AI systems should be
designed, trained, and deployed in ways that remain responsive to
heterogeneous and uncertain lived realities. The ecological emphasis
here does not mean infrastructure availability or interface usability
alone; it concerns the fit between cognitive strategies, available
information, institutional constraints, and the environments in which
decisions are actually made \citep{todd2012,ji2026authoritarian}. Ecological
rationality asks whether the system supports the social, cultural,
institutional, and interpretive conditions through which people can
understand, test, revise, and contest possible futures. This imperative
connects to, but goes beyond, participatory and value-sensitive
approaches alone. Participatory AI research rightly emphasizes that affected
communities should help shape the goals, assumptions, and uses of AI
systems; critical work also warns that participation can become
extractive or merely legitimating when it fails to redistribute
authority \citep{sloane2022,birhane2022,delgado2023}. Ecological rationality takes this concern into consequential, aspirational
contexts. The question is not just whether affected persons were
consulted or diverse stakeholders included, but whether the system helps sustain the conditions under which persons and groups can deliberate
about futures that remain uncertain, socially fragile, and still open to revision.

Ecological rationality matters as the rationality of an AI
recommendation depends on the situation in which it is taken up: the
same recommendation may have different meanings, constraints, and
consequences across contexts, some of which become high-stakes and
difficult to reverse only over time. For a well-supported user, a career
or medical recommendation may be one input among many. For someone with
fewer trusted guides, weaker institutional literacy, and little room for
error, it may become the dominant map of the future. Ecological
rationality therefore requires systems to avoid assuming stable
preferences, equal literacy, reversible options, or equal ability to
absorb mistakes, and to train and evaluate practices that not only
optimize for dominant pathways of success, user satisfaction, or
short-term compliance. Positive alignment and pluralistic value work are
useful insofar as they emphasize context-sensitive, user-authored, and
pluralistic flourishing \citep{laukkonen2026,sorensen2024},
but ecological rationality adds that ``user-authored'' futures require
social conditions in which authorship and agency can cultivate and
develop, without pretending that such agency formation is unconstrained,
fully transparent, or easily achieved. In disadvantaged contexts, where
margins for experimentation are thin and the cost of misdirection is
high, ecological rationality is not a design supplement. It is an
uncompromised condition for both protecting and enabling life- course determining meta-capacity.

\subsection{Counterfactual Reparation}

The third imperative is counterfactual reparation. AI-mediated
consequential decisions should be governed with attention to the futures
they help foreclose, not only the outcomes they directly produce. When a
system ranks, recommends, classifies, or scripts a path, it can redirect
attention, effort, resources, and self-understanding, reshaping what
actions remain available or thinkable under particular social conditions
\citep{davis2020,davis2023}. If the path later proves harmful, misleading, or
uninhabitable, the damage is not limited to a discrete bad decision. The
person or group may have lost time, alternative opportunities,
confidence, relationships, institutional standing, or the ability to
return to paths that once remained open. The recourse literature already
asks how affected persons or groups can respond to algorithmic decisions,
especially through counterfactual explanation, actionable change,
feasible intervention, and robust reversal of adverse classifications
\citep{wachter2018,ustun2019,venkatasubramanian2020,karimi2021}.
Recent work on algorithmic reparation, redress, and repair further
emphasizes that AI governance must address the aftermath of harm, not
only prevention or prediction \citep{davis2021algorithmic,wong2025,hopkins2025,pi2025,davis2025repair}.

Counterfactual reparation builds from this literature but shifts the
object. In aspirational decisions, the harm is that a contingent
sociotechnical path may have reorganized what the person or group came
to see as possible, worthwhile, or recoverable. Counterfactual
reparation therefore asks industry and institutions to preserve and
repair foreclosed alternatives. This includes durable records of the
assumptions and value frames behind recommendations; opportunities to
revisit AI-mediated decisions; second chances when system-shaped
pathways fail; institutional accountability for overconfident or
misleading guidance; and support for re-entering paths displaced by
earlier recommendations. It also requires design choices that keep
alternatives visible before irreversible commitment occurs.

\subsection{From Better Decisions to Conditions for Choosing}

The three imperatives share a common orientation: they treat AI ethics
as the protection and cultivation of meta-capacity crucial to life-course or life-long choice horizon. Existential honesty
resists the false certainty through which AI turns value-forming futures
into apparently settled choices. Ecological rationality supports the
communal, sociocultural, and institutional conditions through which
consequential decisions can be deliberated, revised, and situated.
Counterfactual reparation responds when AI-mediated pathways have
already foreclosed alternatives, redirected action, or disrupted
possible futures. Together, these imperatives address the illusion of
opting at the level where it arises: the formation of future-directed
agency. An AI system may appear accurate, fair, explainable, supervised
while still weakening the conditions under which persons and groups
become capable of making consequential choices. The normative question in consequential
contexts is therefore whether the system preserves the openness,
uncertainty, contestability, and revisability through which a future can
become genuinely one's own.

In particular, one shared aim of these imperatives is to prevent AI
systems and deploying institutions from shifting the costs of conversion
and drift onto those least able to absorb them. This is especially
decisive for disadvantaged populations. Ethical AI cannot only offer
``better'' responses or ``optimal'' recommendations within already
constrained pathways. It must help protect and cultivate the conditions
under which possible futures can be questioned, compared, revised, and
repaired. Otherwise, AI may preserve the outward signs of choice, while
weakening the meta-capacity needed to make a consequential future genuinely one's own.

\section{Conclusion}

This paper has argued that AI-mediated consequential decisions raise an
ethical problem that cannot be captured by the familiar question of
whether systems make accurate, fair, explainable, or preference-satisfying
recommendations. In many domains, individuals and groups do not approach AI
systems with stable preferences, complete information, and reversible
options. They approach them amid uncertainty about who they may become,
which futures are worth pursuing, and what losses a given path may carry.
In such settings, AI does not only assist information or a decision. It can participate in forming the practical, temporal horizon within which decisions acquire high stakes for life trajectories.

The central contribution of the paper is to shift the normative focus
from better decision outputs to the conditions for consequential choice that matters in life course. The illusion of opting is ethically distinctive because it preserves the outward form of meaningful choice while critically weakening the agency required for genuinely consequential decision-making. Drawing on
Ullmann-Margalit's account of opting, we have characterized consequential
choice as transformative, partly irreversible, and shadowed by foreclosed
alternatives. We then show why AI intensifies the illusion of opting:
It can misrecognize simulated confidence as warranted judgment under radical uncertainty, convert open-ended life questions into pathways that appear optimized while prematurely foreclosing alternatives, and relocate
the burden of misdirection onto users who may have little communal or institutional
support for revising or recovering from a path once taken. In other words, AI-intensified illusion of opting can make a path\textit{ feel chosen} even when the
conditions for forming, testing, contesting, and revising that choice are
thin.

This reframing emphasizes why disadvantaged populations require
special attention in facing the AI-driven illusion of opting. For
users with abundant resources and support, AI-mediated guidance can remain one input
among many: a suggestion to be compared with advice from family,
professionals, institutions, and trusted communities. For users with
fewer reserves, weaker institutional literacy, narrower margins for
experimentation, or limited access to alternative forms of guidance, the
same recommendation may acquire much greater practical and life stakes. A
misdirected educational, employment, medical, legal, or migration-related
path may consume scarce time, confidence, money, relationships, or
eligibility for other opportunities. In these cases, the harm is not only
a poor outcome. It is the erosion of meta-capacity: the crucial agentive capacity to form, question, revise, and aspire under real socioeconomic constraints.

The three imperatives proposed in this paper respond to that erosion.
First, existential honesty requires AI systems and deploying institutions to
acknowledge when a decision concerns an unknowable future rather than a
well-specified optimization problem. Second, ecological rationality requires AI
guidance to be situated in the user's social, cultural, institutional,
and material conditions, rather than treating all users as equally able
to interpret, contest, or absorb the risks of recommendation.
Third, counterfactual reparation requires attention to the alternatives that AI-
mediated pathways help foreclose, including mechanisms for revisiting,
repairing, and re-entering paths displaced by earlier guidance. Together,
these imperatives move AI ethics beyond the procedural assurance that a
choice was presented, consented to, or explained. They ask whether the system preserves the conditions under which options and alternatives remain possible for people living under heterogeneous life conditions.

Broadly, our perspective has implications for both AI evaluation and AI governance.
Evaluation should not only ask whether an AI system delivers accurate
predictions, satisfying recommendations, or locally persuasive
explanations. It should also ask whether the system narrows or expands
the user's capacity to deliberate over ends, compare alternatives, detect
uncertainty, seek human support, and revise a path over time. Governance
should likewise treat consequential AI systems as participants in
trajectory formation, not only as tools that issue discrete outputs.
This requires institutional duties around uncertainty disclosure,
escalation to human and community support, preservation of alternatives,
auditability of value assumptions, and repair when system-shaped pathways
become harmful or uninhabitable.

As AI systems move towards shaping education, work, health,
migration, legal access, and personal and collective development, ethical AI in consequential-decision domains must therefore focus on more than short-term choice architecture and immediate preference satisfaction or behavioral support. It must protect the fragile
conditions through which possible futures of life remain open, contestable,
revisable, and repairable. Without systems that make uncertainty explicit, situate guidance in lived constraints, and keep repair possible, AI may preserve the surface grammar of freedom and autonomy while hollowing out the deeper agency needed for consequential choice and aspirational life-direction.

\bibliography{references}
\end{document}